\pdfoutput=1

\documentclass[11pt]{article}

\usepackage{authblk}

\usepackage{acl}

\usepackage{times}
\usepackage{latexsym}
\usepackage{multirow}
\usepackage{array}
\usepackage[T1]{fontenc}

\usepackage[utf8]{inputenc}

\usepackage{microtype}
\usepackage{CJKutf8}
\usepackage{ulem}
\title{Overview of CTC 2021: Chinese Text Correction for Native Speakers}

\author[1]{\bf{Honghong Zhao}}
\author[1,2]{\bf{Baoxin Wang}}
\author[1]{\bf{Dayong Wu}}
\author[2]{\bf{Wanxiang Che}}
\author[1]{\bf{Zhigang Chen}}
\author[1]{\bf{Shijin Wang}}
\affil[1]{State Key Laboratory of Cognitive Intelligence, iFLYTEK Research, China}
\affil[2]{Research Center for SCIR, Harbin Institute of Technology, Harbin, China}
\affil[ ]{\texttt {\{hhzhao4,bxwang2,dywu2,zgchen,sjwang3\}@iflytek.com}}
\affil[ ]{\texttt {car@ir.hit.edu.cn}}

\begin{document}
\normalem
\maketitle
\begin{abstract}

In this paper, we present an overview of the CTC 2021, a Chinese text correction task  for native speakers. We give detailed descriptions of the task definition and the data for training as well as evaluation. We also summarize the approaches investigated by the participants of this task. We hope the data sets collected and annotated for this task can facilitate and expedite future development in this research area. Therefore, the pseudo training data, gold standards validation data, and entire leaderboard is publicly available online at \url{https://destwang.github.io/CTC2021-explorer/}.

\end{abstract}

\section{Introduction}

Chinese text correction (CTC)	is a challenging task in natural language processing and it has attracted more and more concerns recently. We organized the CTC competition in 2021, with a focus on text errors produced by native Chinese speakers. In particular, our task is defined as to detect various errors in the text from native speakers and return the corrected texts.

The previous research on text errors in Chinese is mainly studied using texts written by Chinese-as-a-second-language (CSL) learners, with a focus on grammatical errors or spelling errors \cite{wu2013chinese,yu2014overview,tseng2015introduction,zhao2018overview,rao2020overview,wang2021dynamic}. However, most of the errors in Chinese learner texts seldom appear in texts written by native speakers. Therefore, CTC 2021 collects texts written by native Chinese speakers from the Internet to evaluate the performance of the trained models. These texts are more complex, and the errors are more diverse, including spelling error, grammatical error, and Chinese semantic error. In addition, the corresponding corrections are more rigorous.

The goal of the task is to develop techniques to automatically detect and correct errors made by native Chinese speakers. We provide large-scale pseudo training data, and we release the validation data in which errors have been annotated by native speakers. Blind testing data is used to evaluate the outputs of the participating teams using a common scoring script and evaluation metric.

A total of 124 teams signed up for the task, 42 of them reached the final, and 20 of them submitted final systems. This overview paper provides detailed descriptions of the task and it is organized as follows. Section 2 gives the task definition. Section 3 presents a detailed introduction of the data sets and our pseudo data construction method. Section 4 provides the evaluation metric and Section 5 reports the results of the participants’ approaches. Conclusions are finally drawn in Section 6.

\section{Task Description}

CTC 2021 evaluates Chinese text correction performance on Internet texts written by native Chinese speakers. The participants should detect and correct errors in given texts written by native Chinese speakers. Each character or punctuation mark occupies 1 spot for counting location. The input instance is given a unique passage number “PID”. If the text contains no spelling errors, the checker should return “PID, -1”. If an input text contains at least one error, the output format is “PID [, location, error type, detect word, correct word] +”, where the symbol “+” indicates there is one or more instance of the predicting element “[, location, error type, incorrect word, correct word]”. “Location” denotes the start location of incorrect word. “Error type” can be divided into three coarse categories and seven fine-grained categories, shown in Table~\ref{tab:error-categories}. “Incorrect word” and “correct word” respectively denote the continuous incorrect characters and its correct version. Table~\ref{tab:error-examples} presents some examples. There are two errors in Ex. 1, the $20^{th}$ character "\begin{CJK*}{UTF8}{gbsn}轮\end{CJK*}" should be "\begin{CJK*}{UTF8}{gbsn}论\end{CJK*}", Location “-1” denotes that there is no error in Ex. 2, in Ex. 3 missing a character "\begin{CJK*}{UTF8}{gbsn}供\end{CJK*}" in location 13, and the $26^{th}$ or $27^{th}$ character "\begin{CJK*}{UTF8}{gbsn}都\end{CJK*}" is redundant.

\begin{table}[!tp]
\centering
\begin{tabular}{ll}
\hline
\textbf{coarse categories} & \textbf{fine-grained categories}\\
\hline
\multirow{2}{*}{spelling error} & {character error}\\
& {word error}\\
\hline
\multirow{3}{*}{grammatical error} & {missing error}\\
& {redundant error}\\
& {disordered error}\\
\hline
\multirow{2}{*}{Chinese semantic error} & {semantic repetition}\\
& {syntactic hybridity}\\
\hline
\end{tabular}
\caption{Error type categories.}
\label{tab:error-categories}
\end{table}

\begin{table}[!tb]
\centering
\begin{tabular}{|m{3em}|m{13em}|}
\hline
\multicolumn{2}{|c|}{\textbf{Example 1}}\\
\hline
{input} & \begin{CJK*}{UTF8}{gbsn}{PID=0011-1 关于瑞典时装公司HM拒绝使用新疆产品的言轮在华引发广泛声讨和抵制浪潮，有记者就此提问。华春莹标识：}\end{CJK*}\\
\hline
{output} & \begin{CJK*}{UTF8}{gbsn}{PID=0011-1, 20, character error, 轮, 论, 46, word error, 标识, 表示,}\end{CJK*}\\
\hline
\multicolumn{2}{|c|}{\textbf{Example 2}}\\
\hline
{input} & \begin{CJK*}{UTF8}{gbsn}{PID=0011-2 新疆棉花是世界上最好的棉花之一，不用是相关企业的损失；}\end{CJK*}\\
\hline
{output} & \begin{CJK*}{UTF8}{gbsn}{PID=0011-2, -1}\end{CJK*}\\
\hline
\multicolumn{2}{|c|}{\textbf{Example 3}}\\
\hline
{input} & \begin{CJK*}{UTF8}{gbsn}{PID=0011-3 给老百姓包括少数民族群众提更多的就业机会，一般正常人都都会觉得是件好事。}\end{CJK*}\\
\hline
{output} & \begin{CJK*}{UTF8}{gbsn}{PID=0011-3, 13, missing error, , 供, 27, redundant error, 都, ,}\end{CJK*}\\
\hline
\multicolumn{2}{|c|}{\textbf{Example 4}}\\
\hline
{input} & \begin{CJK*}{UTF8}{gbsn}{PID=0011-4 因为他们自己上历史真的就这么干了上百年，所以现在以己度人；}\end{CJK*}\\
\hline
{output} & \begin{CJK*}{UTF8}{gbsn}{PID=0011-4, 6, disordered error, 上历史, 历史上,}\end{CJK*}\\
\hline
\multicolumn{2}{|c|}{\textbf{Example 5}}\\
\hline
{input} & \begin{CJK*}{UTF8}{gbsn}{PID=0023-1 对学校的未来发展，专家们提出了许多真知灼见的意见。}\end{CJK*}\\
\hline
{output} & \begin{CJK*}{UTF8}{gbsn}{PID=0023-1, 21, semantic repetition, 的意见, ,}\end{CJK*}\\
\hline
\multicolumn{2}{|c|}{\textbf{Example 6}}\\
\hline
{input} & \begin{CJK*}{UTF8}{gbsn}{PID=0069-1 高速公路上交通事故的主要原因是司机违反交通规则或操作不当造成的。}\end{CJK*}\\
\hline
{output} & \begin{CJK*}{UTF8}{gbsn}{PID=0069-1, 29, syntactic hybridity, 造成的, ,}\end{CJK*}\\
\hline
\end{tabular}
\caption{Some examples used in our task.}
\label{tab:error-examples}
\end{table}

\section{Data Preparation}

This section presents the pseudo training data, validation data, and testing data in our task. The texts used in our task were collected from the Internet, including education, science, technology, and other types of data, the collected texts were written by native speakers. Table~\ref{tab:dataset} shows statistics for the data set.

\begin{table}[!tb]
\centering
\begin{tabular}{lrrc}
\hline
& \verb|#|\textbf{Texts} & \verb|#|\textbf{ErrText} & \textbf{AvgLen}\\
\hline
{Train} & {217,634} & {217,630} & {53.51}\\
{Valid} & {969} & {480} & {48.93}\\
{Test} & {967} & {466} & {50.63}\\
\hline
\end{tabular}
\caption{The distributions of error types in validation data.}
\label{tab:dataset}
\end{table}

\subsection{Training data}

We randomly select 217,634 texts from collected Internet texts to create pseudo training data. We randomly choose one position or two for error in each text. If the $i^{th}$ word is chosen, (1) replace it with a pinyin similar word, a shape similar word or a random word (2) delete it (3) insert a word (4) swap it with an adjacent word. Then we get texts with incorrect characters and corresponding corrections.

\subsection{Validation and testing data}

The validation data and testing data use pseudo data construction method same as training data. Finally, we select 969 texts as validation data, 967 as testing data. To obtain gold edits of errors, two annotators annotated these text errors. The error types distribution of validation data is shown in Table~\ref{tab:error-number}, the testing data has a similar distribution.

\begin{table}[!tb]
\centering
\begin{tabular}{lc}
\hline
\textbf{error type} & \textbf{error number}\\
\hline
{spelling error} & {280}\\
{grammatical error} & {158}\\
{Chinese semantic error} & {100}\\
\hline
\end{tabular}
\caption{The distributions of error types in validation data.}
\label{tab:error-number}
\end{table}

\section{Performance Metrics}

Table~\ref{tab:confusion-matrix} shows the confusion matrix used for performance evaluation. In the matrix, TP (True Positive) is the number of errors that are correctly identified by the checker; FP (False Positive) is the number of wrongly identified errors that are non-existent; TN (True Negative) is the number of sentences without any errors which are correctly identified as such; FN (False Negative) is the number of errors which are not detected.

\begin{table}[!tp]
\newcommand{\tabincell}[2]{\begin{tabular}{@{}#1@{}}#2\end{tabular}}
\centering
\begin{tabular}{|cc|c|c|}
\hline
\multicolumn{2}{|c|}{\multirow{2}{*}{Confusion Matrix}} & \multicolumn{2}{c|}{\textbf{System Result}}\\
\cline{3-4}
& & {Positive} & {Negative} \\
\hline
\multirow{2}{*}{\tabincell{c|}{\textbf{Gold} \\ \textbf{Standard}}} & {Positive} & {TP} & {FN}\\
\cline{2-4}
& {Negative} & {FP} & {TN}\\
\hline
\end{tabular}
\caption{Confusion matrix for evaluation.}
\label{tab:confusion-matrix}
\end{table}

Correctness is determined at two levels. The error type does not affect the evaluation results.

(1) Detection level: all locations and incorrect characters in a given text should be completely identical with the gold standard. It is noteworthy that there are some errors shown with different location and incorrect character, but the correction text is right, this is also right detection. For example, “PID=0011-3, 26, redundant error, \begin{CJK*}{UTF8}{gbsn}都\end{CJK*}, ,” is completely identical with “PID=0011-3, 27, redundant error, \begin{CJK*}{UTF8}{gbsn}都\end{CJK*}, ,”.

(2) Correction level: all locations, incorrect characters, and corresponding corrections should be completely identical with the gold standard, or the correction text is completely identical with the gold standard.

The following metrics are measured at both levels with the help of the confusion matrix.

	$Precision=TP/(TP+FP)$
	
	$Recall=TP/(TP+FN)$
	
	$F1=2*Precision*Recall/(Precision+Recall)$
	
Finally, we calculate overall F1 by weighted detection level F1 and correction level F1, defined as: 

    $F1_{overall}=0.8*F1_{detection}+0.2* F1_{correction}$.

Take for example, testing input with gold standards and the system may output the result shown in Table~\ref{tab:matric-example}, the evaluation tool will yield the following performance.

\begin{table}[!tp]
\centering
\begin{tabular}{m{19em}}
\hline
\textbf{Testing output with gold standards}\\
\hline
\begin{CJK*}{UTF8}{gbsn}{PID=0011-1, 20, character error, 轮, 论, 46, word error, 标识, 表示,}\end{CJK*}\\
\begin{CJK*}{UTF8}{gbsn}{PID=0011-2, -1}\end{CJK*}\\
\begin{CJK*}{UTF8}{gbsn}{PID=0011-3, 13, missing error, , 供, 27, redundant error, 都, ,}\end{CJK*}\\
\begin{CJK*}{UTF8}{gbsn}{PID=0011-4, 6, disordered error, 上历史, 历史上,}\end{CJK*}\\
\begin{CJK*}{UTF8}{gbsn}{PID=0023-1, 21, semantic repetition, 的意见, ,}\end{CJK*}\\
\begin{CJK*}{UTF8}{gbsn}{PID=0069-1, 29, syntactic hybridity, 造成的, ,}\end{CJK*}\\
\hline
\textbf{System output result}\\
\hline
\begin{CJK*}{UTF8}{gbsn}{PID=0011-1, \dashuline{20, character error, 轮}, 语,}\end{CJK*}\\
\begin{CJK*}{UTF8}{gbsn}{PID=0011-2, -1}\end{CJK*}\\
\begin{CJK*}{UTF8}{gbsn}{PID=0011-3, \uwave{26, redundant error, 都, ,} 32, character error, 件, 个,}\end{CJK*}\\
\begin{CJK*}{UTF8}{gbsn}{PID=0011-4, 6, redundant error, 上, ,}\end{CJK*}\\
\begin{CJK*}{UTF8}{gbsn}{PID=0023-1, -1}\end{CJK*}\\
\begin{CJK*}{UTF8}{gbsn}{PID=0069-1, \uwave{29, syntactic hybridity, 造成的, ,}}\end{CJK*}\\
\hline
\end{tabular}
\caption{Testing input with gold standards and System output result. Dashed underline indicates detection is right but correction is wrong, wavy underline indicates both detection and correction are right.}
\label{tab:matric-example}
\end{table}

\textbf{Detection-level:}

$Pre.=0.6(=3/5)$ 

$Rec.=0.4286(=3/7)$

$F1=0.5(=2*0.6*0.4286/(0.6+0.4285))$

\textbf{Correction-level:}

$Pre.=0.4(=2/5)$

$Rec.=0.2857(=2/7)$

$F1=0.3333(=2*0.4*0.2857/(0.4+0.2857))$

\textbf{Overall:}

$F1=0.4667(=0.8*0.5+0.2*0.333)$

\section{Evaluation Results}

A total of 124 teams signed up for the task, 42 of them reached the final, and 20 of them submitted final systems. The 20 submitting system teams are all from universities and industries in China. In the official testing phase, each participating team was allowed to submit at most three runs that adopt different models or parameter settings, the highest score of the three times will be the final score of the team. 

In total, we had received 36 runs. Of the 20 submission teams, 16 team systems can work normally. Table~\ref{tab:submission-results} shows the submission statistics for the 16 work normally teams and their final score. we can see that Chinese text correction for native speakers is a challenging task. There remain large gaps between submitting systems and gold standards. In detail, \verb|S&A| gets the best detection F1 score of 0.6800 and best correction F1 score of 0.6460.

\begin{table*}
\centering
\begin{tabular}{ccccccccc}
\hline
\textbf{Team} & \verb|#|\textbf{Runs} & \multicolumn{3}{c}{\textbf{Detection level}} & \multicolumn{3}{c}{\textbf{correction level}} & \textbf{Overall F1}\\
& & {Pre} & {Rec} & {F1} & {Pre} & {Rec} & {F1} & \\
\hline
S\verb|&|A & {3} & 0.6869 & \textbf{0.6733} & \textbf{0.6800} & \textbf{0.6525} & \textbf{0.6396} & \textbf{0.6460} & \textbf{0.6732}\\
\begin{CJK*}{UTF8}{gbsn}{改的都队}\end{CJK*} & {3} & \textbf{0.6890} & 0.5703 & 0.6241 & 0.6316 & 0.5228 & 0.5720 & 0.6137\\
znv\verb|_|sentosa & {1} & 0.4900 & 0.6277 & 0.5503 & 0.3833 & 0.4911 & 0.4306 & 0.5264\\
C\verb|&|L & {3} & 0.5927 & 0.4495 & 0.5113 & 0.5640 & 0.4277 & 0.4865 & 0.5063\\
{MDatai} & {3} & 0.5584 & 0.4733 & 0.5123 & 0.5164 & 0.4376 & 0.4737 & 0.5046\\
{YCC} & {3} & 0.4932 & 0.5030 & 0.4980 & 0.4233 & 0.4317 & 0.4275 & 0.4839\\
{NJU-NLP} & {3} & 0.5448 & 0.4455 & 0.4902 & 0.4407 & 0.3604 & 0.3965 & 0.4715\\
\begin{CJK*}{UTF8}{gbsn}{四条人}\end{CJK*} & {3} & 0.5361 & 0.3386 & 0.4150 & 0.4608 & 0.2911 & 0.3568 & 0.4034\\
ai\begin{CJK*}{UTF8}{gbsn}{编程的小拓}\end{CJK*} & {2} & 0.4648 & 0.3267 & 0.3837 & 0.3831 & 0.2693 & 0.3163 & 0.3702\\
{zybank} & {1} & 0.4579 & 0.3228 & 0.3786 & 0.4017 & 0.2832 & 0.3322 & 0.3693\\
\begin{CJK*}{UTF8}{gbsn}{华夏-龙盈战队}\end{CJK*} & {2} & 0.2550 & 0.3267 & 0.2865 & 0.1947 & 0.2495 & 0.2188 & 0.2730\\
yl\verb|_|test & {1} & 0.4608 & 0.1861 & 0.2652 & 0.2941 & 0.1188 & 0.1693 & 0.2460\\
\begin{CJK*}{UTF8}{gbsn}{晓梦}\end{CJK*} & {1} & 0.3113 & 0.1584 & 0.2100 & 0.2101 & 0.1069 & 0.1417 & 0.1963\\
{only-one} & {1} & 0.365 & 0.1446 & 0.2071 & 0.2550 & 0.1010 & 0.1447 & 0.1946\\
zndx\begin{CJK*}{UTF8}{gbsn}{纠错好难}\end{CJK*} & {1} & 0.3179 & 0.1228 & 0.1771 & 0.1744 & 0.0673 & 0.0971 & 0.1611\\
CTC 2021 baseline & {1} & 0.3361 & 0.0792 & 0.1282 & 0.1849 & 0.0436 & 0.0705 & 0.1167\\
{DAWN} & {1} & 0.0384 & 0.1802 & 0.0633 & 0.019 & 0.0891 & 0.0313 & 0.0569\\
\hline
\end{tabular}
\caption{Submission statistics and results of CTC 2021 in detection level and correction level. CTC 2021 baseline is produced with GECToR, and trained on our CTC 2021 pseudo training data.}
\label{tab:submission-results}
\end{table*}

\begin{table*}[!tp]
\newcommand{\tabincell}[2]{\begin{tabular}{@{}#1@{}}#2\end{tabular}}
\centering
\begin{tabular}{lll}
\hline
\textbf{Team} & \textbf{Approach} & \textbf{Linguistic Resources}\\
\hline
S\verb|&|A & \tabincell{l}{Bert+sequence tagging \\ GECToR \\ Rule-based models} & \tabincell{l}{WeiXin public corpus \\ NLP Chinese corpus \\ Wikimedia \\ SIGHAN \\ Lang-8 \\ Dynamic Corpus of HSK}\\
\hline
\begin{CJK*}{UTF8}{gbsn}{改的都队}\end{CJK*} & \tabincell{l}{GECToR \\ Transformers \\ ReaLise \\ Rule-based models} & \tabincell{l}{WuDaoCorpora 2.0 \\ Wikimedia}\\
\hline
{znv\verb|_|sentosa} & \tabincell{l}{GECToR} & \tabincell{l}{CTC2021 \\ Idiom Example Sentences \\ College Entrance Examination Error Sentences}\\
\hline
{C\verb|&|L} & \tabincell{l}{BERT+CRF+MLM} & \tabincell{l}{SIGHAN \\ HybridSet (\citealp{wang2018hybrid}) \\ XinHua News Corpus \\ CTC2021}\\
\hline
{MDatai} & \tabincell{l}{Electra+sequence tagging \\ ReaLiSe \\ RoBERTa MLM} & \tabincell{l}{CGED \\ NLPCC2018 \\ SIGHAN \\ People's Daily Corpus}\\
\hline
{YCC} & \tabincell{l}{ReaLiSe \\ BERT+CRF+MLM \\ Rule-based models} & \tabincell{l}{SIGHAN \\ CTC2021}\\
\hline
\end{tabular}
\caption{A summary of participants’ developed systems.}
\label{tab:submission-summary}
\end{table*}

We collected the top 6 participants’ reports of their systems. Table~\ref{tab:submission-summary} summarizes their approaches and the usage of linguistic resources for this task. We can observe that most of the participants adopt pre-trained language models (e.g. \citealp{devlin2018bert}; \citealp{liu2019roberta}; \citealp{clark2020electra}), sequence tagging (\citealp{omelianchuk2020gector}), multimodal information of the Chinese characters (\citealp{xu2021read}), sequence to sequence models (\citealp{vaswani2017attention}), and rule-based models. In addition to the CTC 2021 pseudo training data, some public linguistic resources are used by participants, such as Weixin corpus\footnote{\url{https://github.com/nonamestreet/weixin_public_corpus}}, Wikimedia, SIGHAN, NLPCC2018, Lang-8, and WuDaoCorpora 2.0\footnote{\url{https://resource.wudaoai.cn/home}}.

\section{Conclusion}

This paper provides an overview of CTC 2021, including task design, data preparation, evaluation metrics, and evaluation results. The final results show that it is still a challenging task which deserves more concern. In order to provide a good communication platform for researchers, industrial practitioners and NLP enthusiasts, we create CTC2021 leaderboard, release the pseudo training data and gold standards validation data. We hope this task can facilitate and expedite future development in this research area.

\section*{Acknowledgments}
CTC 2021 was hosted by the Chinese Association for Artificial Intelligence, sponsored by iFLYTEK CO.LTD., organized by the Joint Laboratory of HIT and iFLYTEK Research (HFL).

\bibliography{anthology,custom}

\begin{thebibliography}{13}
\expandafter\ifx\csname natexlab\endcsname\relax\def\natexlab#1{#1}\fi

\bibitem[{Clark et~al.(2020)Clark, Luong, Le, and Manning}]{clark2020electra}
Kevin Clark, Minh-Thang Luong, Quoc~V Le, and Christopher~D Manning. 2020.
\newblock Electra: Pre-training text encoders as discriminators rather than
  generators.
\newblock \emph{arXiv preprint arXiv:2003.10555}.

\bibitem[{Devlin et~al.(2018)Devlin, Chang, Lee, and
  Toutanova}]{devlin2018bert}
Jacob Devlin, Ming-Wei Chang, Kenton Lee, and Kristina Toutanova. 2018.
\newblock Bert: Pre-training of deep bidirectional transformers for language
  understanding.
\newblock \emph{arXiv preprint arXiv:1810.04805}.

\bibitem[{Liu et~al.(2019)Liu, Ott, Goyal, Du, Joshi, Chen, Levy, Lewis,
  Zettlemoyer, and Stoyanov}]{liu2019roberta}
Yinhan Liu, Myle Ott, Naman Goyal, Jingfei Du, Mandar Joshi, Danqi Chen, Omer
  Levy, Mike Lewis, Luke Zettlemoyer, and Veselin Stoyanov. 2019.
\newblock Roberta: A robustly optimized bert pretraining approach.
\newblock \emph{arXiv preprint arXiv:1907.11692}.

\bibitem[{Omelianchuk et~al.(2020)Omelianchuk, Atrasevych, Chernodub, and
  Skurzhanskyi}]{omelianchuk2020gector}
Kostiantyn Omelianchuk, Vitaliy Atrasevych, Artem Chernodub, and Oleksandr
  Skurzhanskyi. 2020.
\newblock Gector--grammatical error correction: tag, not rewrite.
\newblock \emph{arXiv preprint arXiv:2005.12592}.

\bibitem[{Rao et~al.(2020)Rao, Yang, and Zhang}]{rao2020overview}
Gaoqi Rao, Erhong Yang, and Baolin Zhang. 2020.
\newblock Overview of nlptea-2020 shared task for chinese grammatical error
  diagnosis.
\newblock In \emph{Proceedings of the 6th Workshop on Natural Language
  Processing Techniques for Educational Applications}, pages 25--35.

\bibitem[{Tseng et~al.(2015)Tseng, Lee, Chang, and
  Chen}]{tseng2015introduction}
Yuen-Hsien Tseng, Lung-Hao Lee, Li-Ping Chang, and Hsin-Hsi Chen. 2015.
\newblock Introduction to sighan 2015 bake-off for chinese spelling check.
\newblock In \emph{Proceedings of the Eighth SIGHAN Workshop on Chinese
  Language Processing}, pages 32--37.

\bibitem[{Vaswani et~al.(2017)Vaswani, Shazeer, Parmar, Uszkoreit, Jones,
  Gomez, Kaiser, and Polosukhin}]{vaswani2017attention}
Ashish Vaswani, Noam Shazeer, Niki Parmar, Jakob Uszkoreit, Llion Jones,
  Aidan~N Gomez, {\L}ukasz Kaiser, and Illia Polosukhin. 2017.
\newblock Attention is all you need.
\newblock \emph{Advances in neural information processing systems}, 30.

\bibitem[{Wang et~al.(2021)Wang, Che, Wu, Wang, Hu, and Liu}]{wang2021dynamic}
Baoxin Wang, Wanxiang Che, Dayong Wu, Shijin Wang, Guoping Hu, and Ting Liu.
  2021.
\newblock Dynamic connected networks for chinese spelling check.
\newblock In \emph{Findings of the Association for Computational Linguistics:
  ACL-IJCNLP 2021}, pages 2437--2446.

\bibitem[{Wang et~al.(2018)Wang, Song, Li, Han, and Zhang}]{wang2018hybrid}
Dingmin Wang, Yan Song, Jing Li, Jialong Han, and Haisong Zhang. 2018.
\newblock A hybrid approach to automatic corpus generation for chinese spelling
  check.
\newblock In \emph{Proceedings of the 2018 Conference on Empirical Methods in
  Natural Language Processing}, pages 2517--2527.

\bibitem[{Wu et~al.(2013)Wu, Liu, and Lee}]{wu2013chinese}
Shih-Hung Wu, Chao-Lin Liu, and Lung-Hao Lee. 2013.
\newblock Chinese spelling check evaluation at sighan bake-off 2013.
\newblock In \emph{SIGHAN@ IJCNLP}, pages 35--42. Citeseer.

\bibitem[{Xu et~al.(2021)Xu, Li, Zhou, Li, Wang, Cao, Huang, and
  Mao}]{xu2021read}
Heng-Da Xu, Zhongli Li, Qingyu Zhou, Chao Li, Zizhen Wang, Yunbo Cao, Heyan
  Huang, and Xian-Ling Mao. 2021.
\newblock Read, listen, and see: Leveraging multimodal information helps
  chinese spell checking.
\newblock \emph{arXiv preprint arXiv:2105.12306}.

\bibitem[{Yu et~al.(2014)Yu, Lee, Tseng, and Chen}]{yu2014overview}
Liang-Chih Yu, Lung-Hao Lee, Yuen-Hsien Tseng, and Hsin-Hsi Chen. 2014.
\newblock Overview of sighan 2014 bake-off for chinese spelling check.
\newblock In \emph{Proceedings of The Third CIPS-SIGHAN Joint Conference on
  Chinese Language Processing}, pages 126--132.

\bibitem[{Zhao et~al.(2018)Zhao, Jiang, Sun, and Wan}]{zhao2018overview}
Yuanyuan Zhao, Nan Jiang, Weiwei Sun, and Xiaojun Wan. 2018.
\newblock Overview of the nlpcc 2018 shared task: Grammatical error correction.
\newblock In \emph{CCF International Conference on Natural Language Processing
  and Chinese Computing}, pages 439--445. Springer.

\end{thebibliography}
\bibliographystyle{acl_natbib}

\end{document}